\documentclass{article}

\usepackage{caption} 
\usepackage[nonatbib,final]{neurips_2018}

\usepackage[utf8]{inputenc} 
\usepackage[T1]{fontenc}    
\usepackage[colorlinks=true,linkcolor=blue,citecolor=blue]{hyperref}       
\usepackage{url}            
\usepackage{booktabs}       
\usepackage{amsfonts}       
\usepackage{nicefrac}       
\usepackage{microtype}      
\usepackage{graphicx}
\usepackage{helvet}
\usepackage{courier}
\usepackage{amsmath}
\usepackage{amssymb}
\usepackage{tabularx}
\usepackage{varwidth}
\usepackage{xcolor}
\usepackage{graphicx}
\usepackage{bbm}
\usepackage{bm}
\usepackage{colortbl}
\usepackage[clock]{ifsym}
\usepackage{enumitem}
\usepackage{multicol} 
\usepackage[utf8]{inputenc}
\usepackage{booktabs}  
\usepackage{algorithm}
\usepackage{algorithmic}
\usepackage{wrapfig}
\usepackage{amsthm}
\usepackage{multirow}
\usepackage{subfig}
\usepackage{setspace}
\usepackage{pifont}
\usepackage[export]{adjustbox}

\newcolumntype{C}[1]{>{\centering\arraybackslash}m{#1}}
\newcolumntype{R}[1]{>{\raggedright\arraybackslash}m{#1}}

\newcommand{\indep}{\raisebox{0.08em}{\rotatebox[origin=c]{90}{$\models$}}}

\usepackage[font=small,skip=10pt]{caption}
\usepackage{setspace}
\DeclareCaptionFormat{myformat}{#1#2#3\hrulefill}

\title{Generalization bounds and algorithms for estimating the individualized treatment effect of dosage}

\title{Generalization bounds and algorithms for estimating conditional average treatment effect of dosage}

  \author{
  Alexis Bellot$^{1,*}$\hspace{0.5cm} Anish Dhir$^{2,*}$ \hspace{0.5cm} Giulia Prando$^{3}$\\
  $^{1}$Columbia University, U.S.A.\hspace{0.3cm} $^{2}$Imperial College London, U.K. \hspace{0.3cm}$^{3}$Babylon Health, U.K.\\
  \texttt{ab5305@columbia.edu} \hspace{0.5cm} \texttt{ad6013@ic.ac.uk} \\
}

\begin{document}

\maketitle

\begin{abstract}
We investigate the task of estimating the conditional average causal effect of treatment-dosage pairs from a combination of observational data and assumptions on the causal relationships in the underlying system. This has been a longstanding challenge for fields of study such as epidemiology or economics that require a treatment-dosage pair to make decisions but may not be able to run randomized trials to precisely quantify their effect and heterogeneity across individuals. In this paper, we extend (Shalit et al, 2017) to give new bounds on the counterfactual generalization error in the context of a continuous dosage parameter which relies on a different approach to defining counterfactuals and assignment bias adjustment. This result then guides the definition of new learning objectives that can be used to train representation learning algorithms for which we show empirically new state-of-the-art performance results across several benchmark datasets for this problem, including in comparison to doubly-robust estimation methods.
\end{abstract}

\section{Introduction}
This paper deals with the estimation of the effect of different dosages (i.e. continuous-valued treatments) on an outcome of interest from a combination of non-experimental (i.e. observational) data and assumptions on the underlying causal graph \cite{pearl2009causality}. This problem is important for decision-making in applied domains where intervening involves not only the choice of treatment type but also the choice of dosage or amount of treatment to be administered. Examples are frequent in medicine where, for instance in the context of chemotherapy treatment, the consequences of a sub-optimal choice of dosage can range from having only a muted effect on tumor growth to toxic effects on the body, as described by \cite{gurney2002calculate}. In this case, while admissible dosage intervals for medical treatments are often determined from clinical trials, these trials often have a small number of patients and evaluate average effects over a restricted population instead of effects conditional on patient characteristics that account for patient heterogeneity, i.e. differences in effects within a patient population \cite{ursino2017dose}. Fortunately, there is a wealth of observational data available in the medical domain from electronic health records that include wider patient groups. Similar observations can be made for other domains such as economics and sociology, see for example \cite{imbens2000role,imai2004causal}.

As a learning problem, the observed dosage in data may depend on confounding variables associated with its assignment. For instance, not all cancer patients are equally likely to be offered the same chemotherapy regiment. This introduces bias in causal effects without adjusting for confounders and variance in the estimation of counterfactuals due to the systematic differences in the distribution of confounding variables between any two dosage levels. In the literature, this challenge has mostly been studied in the binary treatment regime \cite{zhang2020learning,shalit2017estimating,shi2019adapting,johansson2020generalization} where two natural treatment groups emerge and two distributions of data may be compared to analyse bias in treatment assignment and difference in treatment outcomes. The challenge in the context of a continuous dosage parameter is that we must account for differences in confounding variable distributions over an infinite set of counterfactuals, one for each potential dosage level, instead of over a discrete set of treatment groups. There is no notion of treatment group as each individual gets assigned a potentially different and unique treatment value, and definitions and interpretation of a counterfactual and counterfactual errors can be quite different which makes the theoretical extension from binary to continuous dosages non-trivial. 
Previous works have focused mainly on extending binary treatment neural network architectures \cite{shalit2017estimating} to continuous treatments \cite{schwab2020learning, nie2021vcnet, bica2020estimating}. As theoretical contributions for continuous dosages have focused on estimators of population level treatment effects \cite{shi2019adapting, nie2021vcnet}, there are no theoretically motivated loss functions for the estimation of heterogeneous treatment effects.

This paper revisits regularization techniques with an eye towards extending the family of representation learning-based algorithms for predicting conditional average treatment effects (CATE) in the context of a continuous dosage parameter (first proposed by \cite{shalit2017estimating} in the binary setting). We show that the expected error in counterfactual estimation is bounded by the factual error plus a term that quantifies the statistical dependence between the assigned dosage and confounding variables which includes known bounds as a special case. This bound also suggests a new training objective for this problem that we implement in combination with recently developed neural network architectures \cite{nie2021vcnet,schwab2020learning} and thus generalizes several previous proposals to the problem of predicting the effect of dosage. Our contributions may be summarized as follows:
\begin{enumerate}[leftmargin=*]
    \item \textit{Theoretical bounds.} Following \cite{shalit2017estimating}, we prove a new bound on the expected error in counterfactual estimation over all possible dosage choices that is composed by the factual error plus a term that quantifies the statistical dependence between the assigned dosage and confounding variables.
    \item \textit{Algorithms.} We demonstrate that this bound may guide the definition of new training objectives for which we establish new state-of-the-art performance results on several benchmark datasets used for this problem.
\end{enumerate}
And finally, using these results we make one of the first broad comparisons between representation learning techniques using doubly-robust regularization strategies \cite{shi2019adapting,nie2021vcnet} designed for optimal (population) average causal effects, and methods optimizing for the generalization error of conditional average causal effects (the proposed approach).


\section{Preliminaries}
We start by introducing the notation and definitions used throughout the paper. In particular, we use
capital letters for random variables $(X)$, small letters for their values $(x)$, bold letters for sets of variables $(\mathbf X)$ and their values $(\mathbf x)$, and calligraphic letters for the spaces where they are defined $(\mathcal X)$ if not explicitly stated. To simplify notation, we consistently use the shorthand $p(\mathbf x)$ to represent probabilities or densities $p(\mathbf X = \mathbf x)$. For three sets of variables $\mathbf X,\mathbf Y,\mathbf Z$ the conditional independence statement "$\mathbf X$ is conditionally independent of $\mathbf Y$ given $\mathbf Z=\mathbf z$" is written as $\mathbf X \indep \mathbf Y | \mathbf Z$.

We use the semantics of the Rubin-Neyman potential outcomes framework, see e.g. Section 2 in \cite{rubin2005causal}. We assume that for a unit (or individual) with observed covariates $\mathbf x \in\mathcal X$, and tuple $t=(w,s)$ defining the treatment type out of $k$ distinct treatments $w\in\mathcal W=\{w_1, \dots, w_k\}$ and dosage parameter $s\in\mathbb R$, there is a corresponding potential outcome $Y(t)$ that would have been observed had the assigned treatment been $t$. 

The goal is to derive unbiased estimates of the expected potential outcomes for a given set of input covariates: $\mathbb E[Y (t)|\mathbf x]$, for any value of $t$ and $\mathbf x$. However, with observational data only one of these potential outcomes is observed for each unit depending on the treatment assignment. Let $Y$ denote the observed outcome. We will refer to the unobserved potential outcomes as counterfactuals. Let $\mathcal T = \{(w,s):w\in\mathcal W, s\in\mathbb R\}$ denote the set of all treatment options. Under the following standard assumptions, it is well understood that the treatment effect between two selected treatment options $t_1$ and $t_2$ reduces to a contrast of conditional distributions, presented in Lemma 1 below.

\textbf{Assumption 1} (Unconfoundedness).\textit{ The treatment assignment, $t=(w,s)\in\mathcal T$, and potential outcomes, $Y(t)$, are conditionally independent given the covariates $\mathbf x$, i.e. $Y(t)\indep t | \mathbf x$.}

\textbf{Assumption 2} (Overlap). \textit{For any $\mathbf x \in\mathcal X$ such that $p(\mathbf x) > 0$, we have $1 > p(t|\mathbf x) > 0$ for each $t\in\mathcal T$.}

\textbf{Assumption 3} (Consistency). \textit{The observed outcome is the potential outcome, as a function of treatment, when the treatment is set to the observed exposure, i.e. $Y = Y(t)$ if $T=t$ for any $t \in\mathcal T$.}

\textbf{Lemma 1} (Identifiability of the treatment effect). \textit{Under assumptions 1 and 3, and any $t_1,t_2\in\mathcal T$,}
\begin{align}
    \mathbb E [Y(t_1) - Y(t_2)|x] = \mathbb E [Y|x,t_1] - \mathbb E [Y|x, t_2],
\end{align}
which is composed entirely of observable quantities and can be estimated from data given Assumption 2 on overlap.
We refer to the quantity  $\mathbb E [Y(t_1) - Y(t_2)|x]$ as the conditional (or individual if the conditioning set identifies a unit) treatment effect (CATE), and the $\mathbb E [Y(t_1) - Y(t_2)]$ as the average, or population, treatment effect (ATE).

\subsection{Related work} 
In the literature on treatment effect estimation, a substantial amount of work has considered learning representations that correct for assignment bias, most prominently in the binary treatment case, see e.g. \cite{johansson2016learning,shalit2017estimating,yao2018representation,johansson2020generalization,shi2020invariant,zhang2020learning,curth2021nonparametric}. Prediction in the context of binary treatments involves building a separate regression model for each treatment; an approach that does not generalise to the dosage setting due to the now infinite number of possible treatments available. One attempt, however, was made by \cite{schwab2020learning} with their proposed Dose Response networks (DRNets) that consist of multi-task layers for dosage sub-intervals defined on top of a common representation -- the dosage interval being subdivided into equally sized sub-intervals and a multi-task head added for each sub-interval. As a consequence however, the mapping learned by \cite{schwab2020learning} does not account for heterogeneity within dosage sub-intervals and misses the expectation of continuously varying predictions as a function of dosage. \cite{nie2021vcnet} instead develop a Varying Coefficient network (VCNet) that overcomes the need to define sub-intervals resulting in smooth dose-response curves. Other modelling approaches include generative models such as SCIGAN \cite{bica2020estimating} using GAN architectures and propensity weighting using generalized propensity scores, see e.g. \cite{imbens2000role,imai2004causal}.

Propensity scores are also used to design doubly robust\footnote{Doubly robust, in the sense that estimators are consistent if either the propensity score or the prediction function is well-specified and even if one of them is inconsistent.} estimators for the estimation of \textit{population average} counterfactuals $\mathbb E_{\mathbf X}\mathbb E[Y (t)|\mathbf x]$ (which is typically the target quantity in randomized trials), see e.g. \cite{van2006targeted,chernozhukov2017double} for details. In this line of research, \cite{shi2019adapting} and \cite{nie2021vcnet} for binary and continuous treatments respectively, propose to use a modified training objective to encourage solutions that have non-parametrically optimal asymptotic properties for the average treatment effect. Both, however, rely on (approximately) task-agnostic fitted models for counterfactuals $\mathbb E[Y (t)|\mathbf x]$ without (explicit) consideration for the imbalance between dosage levels or broader generalization guarantees which suggests that optimal average causal effects, $\mathbb E_{\mathbf X}\mathbb E[Y (t)|\mathbf x]$, do not necessarily imply optimal conditional average causal effects, $\mathbb E[Y (t)|\mathbf x]$. 

In this paper, we propose for the first time training objectives with generalization guarantees for the prediction of conditional average treatment effects in the context of a continuously-valued dosage parameter that may be used with any differentiable architecture. This extends previous proposals for this problem but also provides an interesting contrast with doubly-robust methods which target a different quantity that may not necessarily be optimal for more individualized predictions, as we discuss in Section \ref{sec:doubly_robust} and in our experiments.


\section{Representation learning for counterfactual dosage prediction}
This section aims at establishing generalization guarantees and algorithms for the counterfactual estimation of the effect of treatment-dosage pairs.  

\subsection{Preliminary definitions and bounds on counterfactual error}
\label{sec:gen_guarantee}

We will discuss representation functions of the form $\phi : \mathcal X \rightarrow \mathcal R$, where $\mathcal R$ is the representation space, and $h : \mathcal R \times \mathcal T \rightarrow \mathcal Y$ is a prediction function defined over $\mathcal R$. We make the following assumption about $\phi$. 

\textbf{Assumption 4}. \textit{The representation $\phi$ is a twice differentiable, one-to-one function. Without loss of generality we will assume that $\mathcal R$ is the image of $\mathcal X$ under $\phi$. Define $p_{\phi}$ to be the distribution induced by $\phi$ over $\mathcal R$\footnote{The remark has been made that the invertibility of representation is difficult to enforce \cite{zhang2020learning}. An algorithmic solution, discussed in \cite{zhang2020learning}, is to include a decoder from the representation to the input domain and reconstruction loss in the training objective to encourage solutions with invertible latent representations.}.}

Consider an arbitrary measure of error $L : \mathcal Y \times \mathcal Y \rightarrow \mathbb R^+$. We define two complimentary loss functions: one is the standard machine learning loss, which we will call the factual error on the estimation at the observed treatment type and dosage tuple, and the other is the counterfactual error, as an average error over all other treatment assignment options, made on the estimation at a selected treatment type and dosage tuple.

\textbf{Definition 1} (Expected loss). \textit{The expected loss for the unit and treatment tuple $(x, t)$ is:
\begin{align}
    l_{h,\phi}(\mathbf x, t) = \int_{\mathcal Y} L(Y(t), h(\phi(\mathbf x), t))p(Y(t)|\mathbf x)dY(t).
\end{align} 
The expected factual and counterfactual losses of $h$ and $\phi$ at treatment $t\in\mathcal T$ are:
\begin{align}
    \epsilon_F(t) &= \int_{\mathcal X}l_{h,\phi}(\mathbf x, t)p(\mathbf x|t) d\mathbf x,\\
    \epsilon_{CF}(t) &= \mathbb E_{p(t')}\int_{\mathcal X}l_{h,\phi}(\mathbf x, t)p(\mathbf x|t') d\mathbf x = \int_{\mathcal X}l_{h,\phi}(\mathbf x, t)p(\mathbf x) d\mathbf x,
\end{align}
where the last equality follows from the definition of expectations and conditional probability. }

The counterfactual error in this definition has a different interpretation than in the binary treatment case. The counterfactual error defines the error made at a chosen treatment tuples $t=(w,s)$ on all populations of units $p(\mathbf x|t')$ for which the observed treatment $t'=(w',s')$ is different than the tuple $t$. For our analysis, it will be of interest to define the average factual and counterfactual error over all treatment options.

\textbf{Definition 2} (Average losses). \textit{The average factual and counterfactual error over all treatment options are defined,
\begin{align}
    \epsilon_F &= \int_{\mathcal T}\epsilon_F(t)p(t)dt = \int_{\mathcal T}\int_{\mathcal X}l_{h,\phi}(\mathbf x, t)p(\mathbf x,t) d\mathbf xdt,\\
    \epsilon_{CF} &= \int_{\mathcal T}\epsilon_{CF}(t)p(t)dt = \int_{\mathcal T}\mathbb E_{p(t_2)}\int_{\mathcal X}l_{h,\phi}(\mathbf x, t)p(\mathbf x|t_2)p(t) d\mathbf xdt  \nonumber\\
    &= \int_{\mathcal T}\int_{\mathcal X}l_{h,\phi}(\mathbf x, t)p(\mathbf x)p(t) d\mathbf xdt, 
\end{align}
respectively\footnote{The integral "$\int_{\mathcal T}dt$" over the product $\mathcal T = \mathcal W \times \mathcal S$ (where $\mathcal W$ is the set of distinct treatments under consideration and $\mathcal S$ is the interval of the real line where dosages are defined) is used as a shorthand for "$\sum_{\mathcal W}\int_{\mathcal S}ds$".}.}

\textbf{Theorem 1} (Generalization bound for the average counterfactual error). \textit{Assume that the unit expected loss functions $l_{h,\phi}(x, t)/B_{\phi}\in G$ for $B_{\phi}>0$ for all $x\in\mathcal X$ and $t\in\mathcal T$, where $G$ is a family of functions $g:\mathcal R \times \mathcal T \rightarrow \mathbb R$. Then, 
\begin{align}
    \epsilon_{CF} \leq \epsilon_{F} + B_{\phi}\cdot\sup_{g\in G} \Big|\int_{\mathcal T}\int_{\mathcal R} g(\mathbf r,t)\cdot (p_{\phi}(\mathbf r)p_{\phi}(t) - p_{\phi}(\mathbf r,t))d\mathbf rdt \Big|. \label{eq:cf_loss_bound}
\end{align}}

This theorem states that the average counterfactual error is upper-bounded by the factual error plus a term that quantifies the dependence between treatment tuple $T$ and covariates $\mathbf X$. If $G$ is a rich enough space of function, the supremum induces an integral probability metric (IPM) such that it equals zero if and only if the random variables induced by the representation is independent of treatment assignment, $\phi(\mathbf X)\indep T$. For example, if $G$ is the space of functions in a universal reproducing kernel Hilbert space with norm bounded by 1, then the regularization term recovers the Hilbert Schmidt Independence Criterion HSIC$(\phi(\mathbf X),T)$ \cite{gretton2007kernel}. And, for characteristic kernels: HSIC$(\phi(\mathbf X),T)=0$ if and only if $\phi(\mathbf X)\indep T$ \cite{sriperumbudur2011universality}. Alternatively, if $G$ is the space of Lipschitz functions with Lipschitz constant bounded by 1 the supremum is defined as the Wasserstein distance between the joint distributions and the product of its marginals \cite{villani2009optimal}.

Such a term is necessary to characterize generalization error because, while counterfactual estimation is unbiased by Assumption 3, the variance of estimating counterfactual outcomes for dosage values that are not heavily represented in observational data may be high. This contributes to higher generalization error by a factor that effectively measures this imbalance using distributional distances. In words, to improve average error one may reduce variance in estimation by minimizing spurious correlations induced by the observational assignment of treatments and dosages, which are responsible for the under-representation of certain dosage values for some sub-populations in the data. 

These spurious correlations in the context of a continuous dosage parameter take the form of high statistical dependence between random variables which generalizes differences in distributions between treatment groups in the binary treatment case. By setting $\mathcal T = \{0,1\}$, that is each individual has two treatment options and counterfactuals are defined by the outcome had one been assigned the alternative treatment, we recover Lemma 1 from \cite{shalit2017estimating}. We show next a similar argument for deriving generalization bounds for the treatment effect comparing two specific treatment options that may be of interest in some applications. We need first to define the error made on the estimation of a counterfactual contrast for a given pair of treatments. 

\textbf{Definition 3} (Treatment effect and error for selected treatment pairs $t_1$ and $t_2$). \textit{Define the treatment effect between two different treatments tuples $t_1, t_2\in\mathcal T$ as,
\begin{align}
    \tau_{(t_1,t_2)}(\mathbf x) = m(t_1,\mathbf x) - m(t_2,\mathbf x),
\end{align}
where we have written $m(t,\mathbf x):= \mathbb E[Y(t)|\mathbf x]$. Define its estimate given a prediction function $f$ by,
\begin{align}
   \hat\tau_{(t_1,t_2)}(\mathbf x) = f(t_1,\mathbf x) - f(t_2,\mathbf x). 
\end{align}
The error in treatment effect estimation is then defined as,
\begin{align}
    \epsilon_{(t_1,t_2)}(f) := \int_{\mathcal X}(\hat\tau_{(t_1,t_2)}(\mathbf x) - \tau_{(t_1,t_2)}(\mathbf x))^2p(\mathbf x)dx.
\end{align}}

\textbf{Theorem 2} (Generalization bound for selected treatment tuples $t_1$ and $t_2$). \textit{With the notation introduced above, it holds that,
\begin{align}
    \epsilon_{(t_1,t_2)}(f)/2 \leq  &\epsilon_{F}(t_1) + \sup_{g\in G} \Big|\int_{\mathcal R} g(\mathbf r)\cdot (p_{\phi}(\mathbf r) - p_{\phi}(\mathbf r|t_1))d\mathbf r \Big| +\epsilon_{F}(t_2)  \nonumber\\ 
    &+ \sup_{g\in G} \Big|\int_{\mathcal R} g(\mathbf r)\cdot (p_{\phi}(\mathbf r) - p_{\phi}(\mathbf r|t_2))d\mathbf r \Big|- \sigma_{Y(t_1)} - \sigma_{Y(t_2)},
\end{align}
where $\sigma_{Y(t_1)}$ and $\sigma_{Y(t_2)}$ stand for the variance of the random variables $Y(t_1)$ and $Y(t_2)$, respectively, under the distribution $p(\mathbf x)$.}


\subsection{Algorithms inspired by bounds on generalization error}
\label{sec:models}
This section discusses the architectures of the representation and prediction functions used, as well as training objectives to leverage the generalization bound in Theorem 1.

The training objective that we define in this work can be used with any neural network architecture that parametrizes a representation and a separate prediction function.  
In this paper, we mainly use architectures of types introduced in  \cite{shalit2017estimating,schwab2020learning,yao2018representation,zhang2020learning}.
These include a representation function $\phi_{\eta}: \mathcal{X} \to \mathcal{R}$ and a corresponding prediction function $h_{\theta}: \mathcal{R}\times \mathcal T \to \mathbb R$, parameterized by sets of parameters $\theta$ and $\eta$ respectively.
These architectures have inductive biases that allow the prediction function to take any changes in treatment types or dosage levels into account.
Without this, the effects of any changes in the treatment or dosage can be diluted in the high dimensional representation. In \cite{shalit2017estimating}, each treatment type $w \in\mathcal W$ has a separate prediction function  $h^{(w)}_\theta:\mathcal{R}\times \mathbb R \to \mathbb R$. 
We use the following architectures that extend this for continuous dosages. Dose Response Networks (DRNet) \cite{schwab2020learning} divide the dosage values into bins with a separate prediction function for each bin.
To allow for more accurate learning of continuous response functions, Varying Coefficient Networks (VCNet) \cite{nie2021vcnet} parametrize the parameters of the prediction head as functions of the dosage $\theta(s)=(\theta_1(s), \dots, \theta_{d_\theta}(s))$, where $d_{\theta}$ is the total number of parameters.
Each scalar parameter is approximated by linear combinations $\theta_i(s)=\sum_{l=1}^L \alpha_{i,l}\psi_l(s)$ of basis polynomial functions $\{\psi_l\}_{l=1}^L$ defined on the space of dosage values $\mathcal S$. The coefficients $\{\alpha_{i,l}:i=1,\dots,d_{\theta}, l=1,\dots, L \}$ are to be optimized such that each prediction function defines a trainable map,
\begin{align}
    h^{(w)}_{\theta}(\mathbf z, s) := h^{(w)}_{\theta(s)}(\mathbf z, s).
\end{align}

Our algorithmic contribution is to explicitly account and adjust for the bias induced by treatment group and dosage level dependence on covariates. Following the discussion in section \ref{sec:gen_guarantee}, we wish to learn a representation $\phi$ and prediction function $h$ that minimize a trade-off between predictive accuracy and imbalance in the representation space using the following objective: 
\begin{align}
    \label{objective}
    \underset{\theta, \eta}{\text{min}} \hspace{0.2cm}\sum_{i=1}^n (y_i - h_{\theta}(\phi_{\eta}(\mathbf x_i),t_i))^2 + \gamma\cdot \text{IPM} (\{\phi_{\eta}(\mathbf{x}_i),\{t_j\}_{j : t_j \neq t_i} \}, \{ \phi_{\eta}(\mathbf{x}_i), t_i \} ),
\end{align}

\begin{figure*}[t]
\captionsetup{skip=5pt}
\centering
\includegraphics[width=0.5\textwidth]{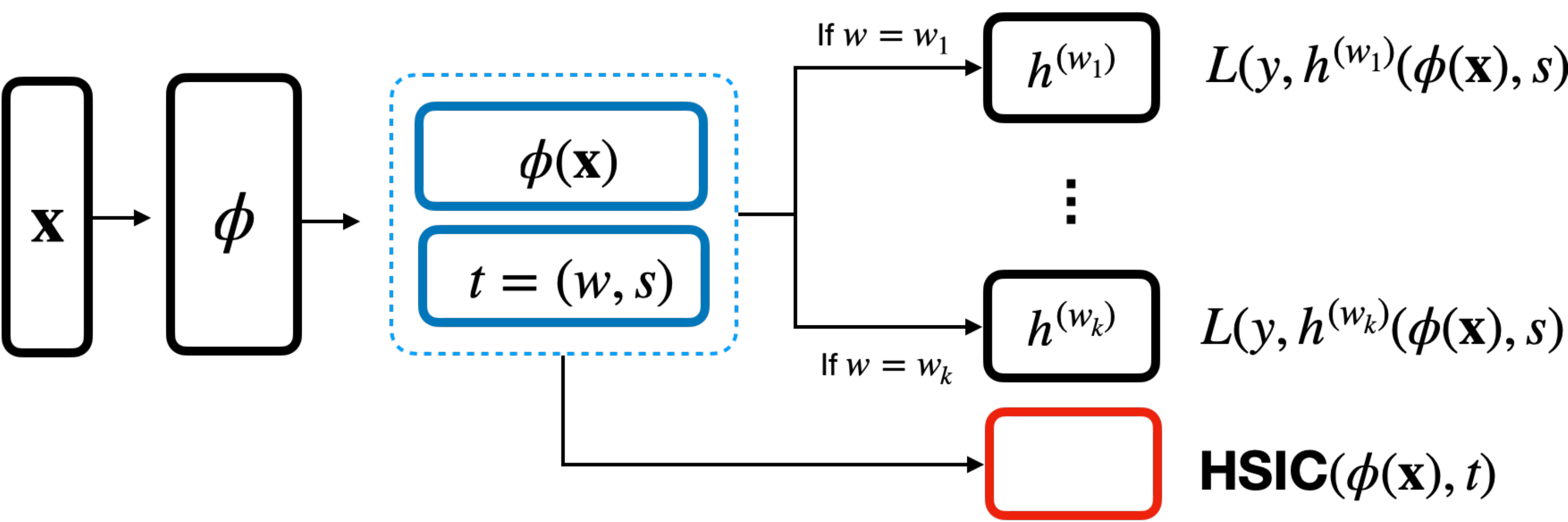}
    \caption{Sketch of the architecture using VCNet and HSIC as the IPM.}
    \label{fig:architecture}
\end{figure*}

where $\gamma \geq 0$ is a hyperparameter. If we choose HSIC as the IPM, we can directly minimize $\text{HSIC}(\phi_{\eta}(\mathbf{x}_i), t_i)$.
For other metrics like Wasserstein, we can force independence by permuting the observed treatment-dosage pair across individuals. 
The representation and the permuted treatments will then form approximate samples from the factorised distribution of the representation and the treatments.
Minimizing the Wasserstein distance between this and the observed joint is then equivalent to forcing independence between the distribution of the treatments and the representation.

We train our models with this objective using stochastic gradient descent, where we backpropagate the error through both the prediction and representation networks, noting that each sample is used to update only the prediction network corresponding to the observed treatment; for example, an observation $(\mathbf x_i, t_i = (w_i, s_i), y_i)$ is only used to update $h^{(w_i)}$, while all samples are used to update $\phi$. A sketch of this training routine is given in Figure \ref{fig:architecture}.

\subsection{Contrast with doubly robust estimation of the average treatment effect}
\label{sec:doubly_robust}
Theorem 1 suggests that the imbalance in the distribution of $\mathbf X$ across treatment dosage pairs is relevant for the expected generalization error of fitted models. Estimators inspired from semi-parametric literature, known as doubly robust estimators \cite{van2006targeted,chernozhukov2017double}, instead follows a different philosophy to optimizing for average treatment effects (ATE), e.g. $\mathbb E_{\mathbf X}\mathbb E[Y (1)|\mathbf x] - \mathbb E_{\mathbf X}\mathbb E[Y (0)|\mathbf x]$, by constructing a prediction function $h_1:\mathcal X \times \mathbb R \rightarrow \mathbb R$, propensity score function $h_2:\mathcal X \rightarrow \mathbb R$, and perturbation terms $\epsilon$, satisfying the non-parametric estimating equation,
\begin{align}
    \frac{1}{n} \sum_{i=1}^n \eta(y_i, t_i, \mathbf x_i; \hat h_1, \hat h_2, \hat \epsilon) = 0,
    \label{eq:non_para_tr}
\end{align}
where (in the binary treatment case for simplicity),
\begin{align}
    \eta(y_i, t_i, \mathbf x_i; h_1, h_2, \epsilon) =  
    h_1(\mathbf x_i,0) - h_1(\mathbf x_i,0) + \left(\frac{t}{h_2(\mathbf x_i)} - \frac{1 - t_i}{1 - h_2(\mathbf x_i)}\right)\cdot(y_i - h_1(t_i, \mathbf x_i)) - \epsilon.
\end{align}
$h_1(t, \mathbf x)$ is an estimator of $\mathbb E[Y (t)|\mathbf x]$, $h_2(\mathbf x)$ is an estimator of the probability of treatment $P(t|\mathbf x)$ and $\epsilon\in\mathbb R$ is a perturbation terms that is optimized. 
In the literature, a common estimation approach is to rely on (task-agnostic) fitted models $\hat h_1$ and $\hat h_2$, and then choose $\epsilon$ so that this equation is satisfied.
If $h_1$ and $h_2$ are consistent estimators of the outcome and propensity scores respectively,
as well as satisfy equation \ref{eq:non_para_tr},
the resulting estimator of the ATE will have desirable asymptotic properties \cite{shi2019adapting, kennedy2016semiparametric}.
However, as these guarantees are on the average treatment effects, they do not necessarily guarantee accurate estimates of conditional treatment effects.

In the context of neural networks, \cite{shi2019adapting,nie2021vcnet} propose to learn a joint representation $\phi(\mathbf x)$ that is conducive to both counterfactual $h_1:\mathcal R \times \mathbb R \rightarrow \mathbb R$ and propensity score estimation $h_2:\mathcal R \rightarrow \mathbb R$ by a using a loss function that trades-off the two objectives, e.g.,
\begin{align}
\label{objective_doubly_robust}
    \frac{1}{n}\sum_{i=1}^n (y_i - h_1(\phi(\mathbf x_i),t_i))^2  +\alpha \cdot\text{CrossEntropy}(h_2(\phi(\mathbf x_i)), t_i),
\end{align}
as in \cite[Eq. (1)]{nie2021vcnet} or \cite[Eq. (2.2)]{shi2019adapting}. The motivation is that: "If the average treatment effect is identifiable conditioning on the propensity score [$\dots$] it suffices to adjust for only the information in $\mathbf x$ that is relevant for predicting the treatment", see \cite[Theorem 2.1]{shi2019adapting}. 
Intuitively, the cross entropy term in equation \ref{objective_doubly_robust} encourages the representation to retain information that is predictive of the treatment.
Hence, it encourages the discarding of information that is predictive of the outcome but not the treatment, which is simply noise when predicting the treatment.

Variables that affect the outcome and not treatment are referred to as effect modifiers in the literature, see e.g. \cite{hernan2010causal}.
By definition, the treatment effect varies across different conditioning sets of these effect modifiers.
As effect modifiers are responsible for the heterogeneity of treatment effects, it is necessary to condition on them to obtain accurate conditional treatment effects.
Thus, to compute conditional average or "individualized" treatment effects such representations may be too restrictive because they tend to ignore effect modifiers.

In contrast, our regularizer penalizes the dependence between the representation and the treatment distributions explicitly. 
Loosely speaking we discard covariate information predictive of treatment but outcome information is retained.
Hence, our regularizer should preserve these effect modifiers leading to more accurate estimates of conditional treatment effects.
We conclude that, in general, optimal average treatment effects does not necessarily imply optimal conditional average treatment effects as measured by expected losses in Definitions 1 and 2\footnote{Definitions 1 and 2 also involve averages but makes a head to head comparisons between observed outcomes and predicted outcomes for each individual in the term $l_{h, \phi}(\mathbf{x}, t)$ (which are then averaged across individuals) instead of averaging predicted counterfactuals across the whole population before comparison with average true outcomes across different dosage levels.}. 
We verify this intuition in our experiments.

\begin{table*}[t]
\fontsize{8.5}{10.5}\selectfont
\centering
\begin{tabular}{| p{1.8cm}|C{1.5cm}|C{1.5cm}|C{1.5cm}|C{1.5cm}|C{1.5cm}|C{1.5cm}|  }

 \cline{2-7}
   \multicolumn{1}{c|}{} &   \multicolumn{2}{c|}{\textbf{Synthetic} }&\multicolumn{2}{c|}{\textbf{IHDP-continuous} }&\multicolumn{2}{c|}{\textbf{News}}\\
 \cline{2-7}
   \multicolumn{1}{c|}{} & $\sqrt{\textbf{MISE}}$ & $\sqrt{\textbf{AMSE}}$  &$\sqrt{\textbf{MISE}}$& $\sqrt{\textbf{AMSE}}$ &$\sqrt{\textbf{MISE}}$& $\sqrt{\textbf{AMSE}}$\\
 \hline
 GPS &  2.80 (0.51) & 2.75 (0.52)  & 4.91 (0.87)  & 4.88 (0.88) & -  & - \\
 \hline\hline
MLP &  0.72 (0.09) & 0.62 (0.11)  & 0.74 (0.07) & 0.59 (0.05) & 1.05 (0.07)  & 0.66 (0.15) \\
\hline\hline
DRNet      &     0.34 (0.06) &   0.20 (0.04) &    0.60 (0.06) &  0.42 (0.05) & 0.84 (0.04) & 0.37 (0.09) \\
\hline
DRNet-PS   &     0.43 (0.07) &   0.25 (0.06) &    0.66 (0.15) &  0.43 (0.12) & 0.84 (0.04) & 0.37 (0.09) \\
\hline
DRNet-TR   &  0.41 (0.03) &  0.17 (0.02) & 2.07 (3.54) & 0.68 (0.70) &  0.82 (0.05) &  0.30 (0.10) \\
\hline
\textbf{DRNet-HSIC} &    0.39 (0.06) &     0.22 (0.02)  &    0.52 (0.05) &  0.35 (0.06) & 0.87 (0.04) & 0.44 (0.07)\\
\hline
\textbf{DRNet-Wass} &     0.38 (0.07) &   0.21 (0.02) &    \textbf{0.51 (0.08)} &  0.34 (0.08) & 0.86 (0.04) & 0.43 (0.07) \\
\hline
\hline
VCNet      &     0.33 (0.03) & 0.13 (0.06) &    0.56 (0.09) &  0.33 (0.09) & 0.85 (0.05) & 0.31 (0.11) \\
\hline
VCNet-PS   &     0.62 (0.42) & 0.32 (0.30) &    0.59 (0.11) & \textbf{0.31 (0.12)} &0.98 (0.09) & \textbf{0.25 (0.13)} \\
\hline
VCNet-TR   &     0.42 (0.12) & 0.19 (0.11) &    0.64 (0.39) &  \textbf{0.31 (0.13)} & 0.99 (0.06) & 0.4 (0.08) \\
\hline
\textbf{VCNet-HSIC} &     \textbf{0.28 (0.04)} & \textbf{0.10 (0.03)} &    0.56 (0.09) &  0.33 (0.08) &  0.87 (0.05) & 0.35 (0.11)
\\
\hline
\textbf{VCNet-Wass} &     0.38 (0.03) & 0.11(0.03) &    0.55 (0.10) &  0.33 (0.08) & \textbf{0.81 (0.05)} & 0.29 (0.1)  \\
 \hline
\end{tabular}
\caption{Performance comparisons in terms of MISE and AMSE on Synthetic, IHDP-continuous, and News datasets.}
\label{table:perf}
\end{table*}

\section{Experiments}
We conduct controlled experiments on synthetic and semi-synthetic datasets previously used by \cite{schwab2020learning,bica2020estimating,nie2021vcnet}. Overall, we found that simulation results support our generalization guarantees with different architectures benefiting from the proposed regularization strategy using both the HSIC and Wasserstein distances. Moreover, the empirical comparison with doubly-robust estimation methods highlights the contrast we outlined in Section \ref{sec:doubly_robust} showing that optimization for average treatment effects is not necessarily adequate for conditional average treatment effect estimation.

\textbf{Comparisons} As a baseline for comparison, we evaluate different neural network architectures without regularization that have been used for this problem, including a standard multilayer perceptron (MLP) that optimises the squared error loss objective to learn the weights of the network, 
a standard VCNet \cite{nie2021vcnet}, and DRNet \cite{schwab2020learning}. 
We also consider the two doubly-robust proposals made by \cite{nie2021vcnet, shi2019adapting}: VCNets and DRNets that only optimises for minimum values of equation \ref{objective_doubly_robust} are denoted VCNet-PS, DRNet-PS, and VCNets and DRNets that minimizes both equations \ref{objective_doubly_robust} and \ref{eq:non_para_tr} are written as VCNet-TR, DRNet-TR. Finally, we consider Generalized Propensity Scores (GPS) \cite{imbens2000role,imai2004causal} as a linear alternative. The proposed methods are labeled DRNet-HSIC, DRNet-Wass, VCNet-HSIC, and VCNet-Wass, which combine existing architectures with regularization methods. We include the details of the network and hyperparameter optimisation in Appendix \ref{sec:alg_details}.

\textbf{Datasets.} The nature of the treatment-effects estimation problem does not allow for meaningful evaluation on real-world datasets. This is simply because we never observe a counterfactual for a given unit. There are, however, established synthetic and semi-synthetic datasets (i.e. with constructed outcome functions but covariates from real-world experiments, see details in the Appendix) that have been used by \cite{schwab2020learning, bica2020estimating,nie2021vcnet}. Following these proposals we use,
\begin{itemize}[leftmargin=*,itemsep=0pt]
    \item \textit{Fully synthetic.} A dataset with a total of 6 randomly generated covariates and a single treatment with dosage ranging from 0 to 1 that involve complex functions for both treatment assignment and outcome function, as defined by \cite{nie2021vcnet}.
    \item \textit{IHDP-continuous.} The original semi-synthetic IHDP dataset from \cite{hill2011bayesian} contains binary treatments with 747 observations on 25 covariates. We adapt this dataset to the continuous dosage context by changing the treatment assignment and outcome function. We generate these in a similar way to \cite{nie2021vcnet}.
    \item \textit{News.} The News dataset consists of 3000 randomly sampled news items from the NY Times corpus \cite{newman2008bag}, which was originally introduced as a benchmark in the binary treatment setting. We generate the treatment and outcome in a similar way as \cite{bica2020estimating}.
\end{itemize}

\textbf{Metrics.} For performance comparisons, we consider the Mean Integrated Squared Error (MISE) and Average Mean Squared Error (AMSE) metrics:
\begin{align}
    \text{MISE} &= \frac{1}{N} \frac{1}{|\mathcal{W}|} \sum_{n=1}^N \sum_{w \in \mathcal{W}} \mathbb{E}_S \left[ (y_{n}(w, s) - \hat{y}_{n}(w, s))^2 \right], \nonumber\\
    \text{AMSE} &= \frac{1}{|\mathcal{W}|}\sum_{w \in \mathcal{W}}   \mathbb{E}_S \left[   \bigg(\frac{1}{N}\sum_{n=1}^N (y_n(w,s) - \hat{y}_n(w,s)) \bigg)^2 \right]  ,\nonumber
\end{align}
where $y_n(w,s)$ and $\hat{y}_n(w,s)$ stand for the true and predicted outcome for individual $n$ given treatment-dosage pairs $(w,s)\in\mathcal T$, and $\mathbb E_Sg(s) =\int g(s)p(s)ds$. Intuitively, MISE calculates how well an algorithm is at estimating individual level dose response and accounts for the heterogeneity in treatment response. 
Whereas AMSE calculates the accuracy of the population level dose response.  
As the doubly robust methods get rid of effect modifiers, that are useful for accurate predictions, but have theoretical guarantees for the average treatment effects, we expect these methods to get a better AMSE.
However, as the regularizers proposed in this work provide guarantees on the counterfactual error, we expect models trained with these to achieve a better MISE score. 

\subsection{Testing the effectiveness of regularisation for treatment effect estimation}
We start our experiments by testing the effectiveness of our proposed regulariser by computing performance as a function of $\gamma$, that determines the influence of the independence
\begin{wrapfigure}{r}{8cm}
\vspace{-0.5cm}
\captionsetup{skip=5pt}
\centering
\includegraphics[width=0.4\textwidth]{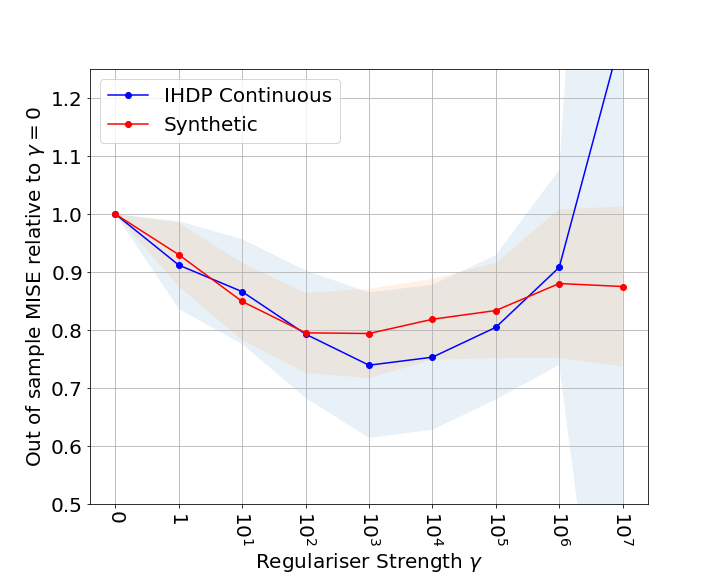}
\caption{Effect of IPM (HSIC) regulariser.}
\label{fig:gamma_vary}
\end{wrapfigure}
constraint in feature space in equation (\ref{objective}). For this experiment we use the VCNET model with the HSIC and compare MISE results for varying values of $\gamma$ relative to $\gamma = 0$ (without regularisation). We see in Figure \ref{fig:gamma_vary} that the presence of treatment assignment bias due to the influence of confounding in both \textit{IHDP continuous} and \textit{Synthetic} datasets confers an advantage to using the proposed regularization term (with increasing $\gamma > 0$) relative to $\gamma = 0$. Across different data generating mechanisms, some value of $\gamma > 0$ gives a significant gain which illustrates our generalization guarantees but also shows that some form of regularisation may broadly be applicable in practice across different data generating mechanisms.

\subsection{Performance comparisons}
Next, in this section we make wide-ranging comparisons against the benchmark prediction algorithms on our 3 datasets. Table \ref{table:perf} reports mean performance and standard deviations on 50 separately generated samples for each of the three datasets\footnote{GPS requires matrix inversion which was not feasible to compute on the high-dimensional News dataset.}. Overall, and especially in terms of MISE, the proposed regularization technique, using either the HSIC or Wasserstein distances between distributions, outperform on average all other regularization techniques on both choices of neural network architecture. Several trends are interesting to discuss in more detail.


It is interesting to see also that, across all datasets, there is a clear trend showing that regularizing for optimal generalization performance in terms of the MISE with the HSIC leads to good population average performance as well, as measured by AMSE. By looking at the performance of methods using regularization inspired by doubly-robust optimality results, we see that the reverse does not always hold: optimizing for optimal population average counterfactuals does not always lead to similarly good individual level performance in terms of MISE. For instance on News, in terms of AMSE, the doubly robust method VCNet-PS ranks among the best performing methods but have only average performance in terms of MISE. On IHDP-continuous as well, the doubly robust methods get the best AMSE (VCNet-PS, VCNET-TR), however their performance on MISE is not anywhere near the best performing model.
In contrast, the HSIC and Wass algorithms get the best MISE as expected but they are not too far off in terms of AMSE performance.
We believe that this discrepancy is due to the doubly robust methods discarding information that helps predict the outcome, resulting in a worse MISE performance.
This also emphasizes the fact that estimating average counterfactuals and individual counterfactuals can require different objectives.

\section{Conclusion}
In this paper, we investigate the task of estimating the conditional average causal effect of dosage from a combination of observational data and assumptions on the causal relationships in the underlying system. When these assumptions hold, we give new bounds on the counterfactual generalization error in the context of a continuous dosage parameter that subsume generalization guarantees from the binary treatment case \cite{shalit2017estimating}. Using this result, we provide new learning objectives that can be used to guide the training of representation learning algorithms. We show empirically new state-of-the-art performance results across several benchmark datasets for this problem. To our knowledge, this is the first paper exploring representation learning and regularization for conditional average counterfactual estimation in the context of continuous-valued treatments in a principled manner. 

\bibliography{bibliography}

\begin{thebibliography}{10}

\bibitem{bica2020estimating}
Ioana Bica, James Jordon, and Mihaela van~der Schaar.
\newblock Estimating the effects of continuous-valued interventions using
  generative adversarial networks.
\newblock {\em Advances in Neural Information Processing Systems},
  33:16434--16445, 2020.

\bibitem{chernozhukov2017double}
Victor Chernozhukov, Denis Chetverikov, Mert Demirer, Esther Duflo, Christian
  Hansen, and Whitney Newey.
\newblock Double/debiased/neyman machine learning of treatment effects.
\newblock {\em American Economic Review}, 107(5):261--65, 2017.

\bibitem{curth2021nonparametric}
Alicia Curth and Mihaela Schaar.
\newblock Nonparametric estimation of heterogeneous treatment effects: From
  theory to learning algorithms.
\newblock In {\em International Conference on Artificial Intelligence and
  Statistics}, pages 1810--1818. PMLR, 2021.

\bibitem{gretton2007kernel}
Arthur Gretton, Kenji Fukumizu, Choon~Hui Teo, Le~Song, Bernhard Sch{\"o}lkopf,
  and Alexander~J Smola.
\newblock A kernel statistical test of independence.
\newblock In {\em NIPS}, 2007.

\bibitem{gurney2002calculate}
Howard Gurney.
\newblock How to calculate the dose of chemotherapy.
\newblock {\em British journal of cancer}, 86(8):1297--1302, 2002.

\bibitem{hernan2010causal}
Miguel~A Hern{\'a}n and James~M Robins.
\newblock Causal inference, 2010.

\bibitem{hill2011bayesian}
Jennifer~L Hill.
\newblock Bayesian nonparametric modeling for causal inference.
\newblock {\em Journal of Computational and Graphical Statistics},
  20(1):217--240, 2011.

\bibitem{hirano2004propensity}
Keisuke Hirano and Guido~W Imbens.
\newblock The propensity score with continuous treatments.
\newblock {\em Applied Bayesian modeling and causal inference from
  incomplete-data perspectives}, 226164:73--84, 2004.

\bibitem{imai2004causal}
Kosuke Imai and David~A Van~Dyk.
\newblock Causal inference with general treatment regimes: Generalizing the
  propensity score.
\newblock {\em Journal of the American Statistical Association},
  99(467):854--866, 2004.

\bibitem{imbens2000role}
Guido~W Imbens.
\newblock The role of the propensity score in estimating dose-response
  functions.
\newblock {\em Biometrika}, 87(3):706--710, 2000.

\bibitem{johansson2016learning}
Fredrik Johansson, Uri Shalit, and David Sontag.
\newblock Learning representations for counterfactual inference.
\newblock In {\em International conference on machine learning}, pages
  3020--3029. PMLR, 2016.

\bibitem{johansson2020generalization}
Fredrik~D Johansson, Uri Shalit, Nathan Kallus, and David Sontag.
\newblock Generalization bounds and representation learning for estimation of
  potential outcomes and causal effects.
\newblock {\em arXiv preprint arXiv:2001.07426}, 2020.

\bibitem{kennedy2016semiparametric}
Edward~H Kennedy.
\newblock Semiparametric theory and empirical processes in causal inference.
\newblock In {\em Statistical causal inferences and their applications in
  public health research}, pages 141--167. Springer, 2016.

\bibitem{kingma2014adam}
Diederik~P Kingma and Jimmy Ba.
\newblock Adam: A method for stochastic optimization.
\newblock {\em arXiv preprint arXiv:1412.6980}, 2014.

\bibitem{newman2008bag}
David Newman.
\newblock Bag of words data set.
\newblock {\em UCI Machine Learning Respository}, 2008.

\bibitem{nie2021vcnet}
Lizhen Nie, Mao Ye, Dan Nicolae, et~al.
\newblock Vcnet and functional targeted regularization for learning causal
  effects of continuous treatments.
\newblock In {\em International Conference on Learning Representations}, 2020.

\bibitem{pearl2009causality}
Judea Pearl.
\newblock {\em Causality}.
\newblock Cambridge university press, 2009.

\bibitem{rubin2005causal}
Donald~B Rubin.
\newblock Causal inference using potential outcomes: Design, modeling,
  decisions.
\newblock {\em Journal of the American Statistical Association},
  100(469):322--331, 2005.

\bibitem{schwab2020learning}
Patrick Schwab, Lorenz Linhardt, Stefan Bauer, Joachim~M Buhmann, and Walter
  Karlen.
\newblock Learning counterfactual representations for estimating individual
  dose-response curves.
\newblock In {\em Proceedings of the AAAI Conference on Artificial
  Intelligence}, volume~34, pages 5612--5619, 2020.

\bibitem{shalit2017estimating}
Uri Shalit, Fredrik~D Johansson, and David Sontag.
\newblock Estimating individual treatment effect: generalization bounds and
  algorithms.
\newblock In {\em International Conference on Machine Learning}, pages
  3076--3085. PMLR, 2017.

\bibitem{shi2019adapting}
Claudia Shi, David Blei, and Victor Veitch.
\newblock Adapting neural networks for the estimation of treatment effects.
\newblock {\em Advances in neural information processing systems}, 32, 2019.

\bibitem{shi2020invariant}
Claudia Shi, Victor Veitch, and David~M Blei.
\newblock Invariant representation learning for treatment effect estimation.
\newblock In {\em Uncertainty in Artificial Intelligence}, pages 1546--1555.
  PMLR, 2021.

\bibitem{sriperumbudur2011universality}
Bharath~K Sriperumbudur, Kenji Fukumizu, and Gert~RG Lanckriet.
\newblock Universality, characteristic kernels and rkhs embedding of measures.
\newblock {\em Journal of Machine Learning Research}, 12(7), 2011.

\bibitem{ursino2017dose}
Moreno Ursino, Sarah Zohar, Frederike Lentz, Corinne Alberti, Tim Friede, Nigel
  Stallard, and Emmanuelle Comets.
\newblock Dose-finding methods for phase i clinical trials using
  pharmacokinetics in small populations.
\newblock {\em Biometrical Journal}, 59(4):804--825, 2017.

\bibitem{van2006targeted}
Mark~J Van Der~Laan and Daniel Rubin.
\newblock Targeted maximum likelihood learning.
\newblock {\em The international journal of biostatistics}, 2(1), 2006.

\bibitem{villani2009optimal}
C{\'e}dric Villani.
\newblock {\em Optimal transport: old and new}, volume 338.
\newblock Springer, 2009.

\bibitem{yao2018representation}
Liuyi Yao, Sheng Li, Yaliang Li, Mengdi Huai, Jing Gao, and Aidong Zhang.
\newblock Representation learning for treatment effect estimation from
  observational data.
\newblock {\em Advances in Neural Information Processing Systems}, 31, 2018.

\bibitem{zhang2020learning}
Yao Zhang, Alexis Bellot, and Mihaela Schaar.
\newblock Learning overlapping representations for the estimation of
  individualized treatment effects.
\newblock In {\em International Conference on Artificial Intelligence and
  Statistics}, pages 1005--1014. PMLR, 2020.

\end{thebibliography}
\bibliographystyle{plain}

\newpage
\appendix
{\Large \textbf{Appendix}}
\\\\
This Appendix is outlined as follows.
\begin{itemize}
    \item Appendix \ref{sec:proofs_app} contains proofs of all theoretical claims.
    \item Appendix \ref{sec:ap_data} contains data generating mechanisms for synthetic and semi-synthetic datasets used in experiments.
    \item Appendix \ref{sec:alg_details} contains details on the implementation and training routines of all algorithms.
\end{itemize}

\section{Proofs}
\label{sec:proofs_app}

\textbf{Theorem 1} (Generalization bound for the average counterfactual error). \textit{Assume that the unit expected loss functions $l_{h,\phi}(x, t)/B_{\phi}\in G$ for $B_{\phi}>0$ for all $x\in\mathcal X$ and $t\in\mathcal T$, where $G$ is a family of functions $g:\mathcal R \times \mathcal T \rightarrow \mathbb R$. Then, 
\begin{align}
    \epsilon_{CF} \leq \epsilon_{F} + B_{\phi}\cdot\sup_{g\in G} \Big|\int_{\mathcal T}\int_{\mathcal R} g(\mathbf r,t)\cdot (p_{\phi}(\mathbf r)p_{\phi}(t) - p_{\phi}(\mathbf r,t))d\mathbf rdt \Big|. \label{eq:cf_loss_bound}
\end{align}}

\textit{Proof.} Let $\psi:\mathcal R \rightarrow \mathcal X$ be the inverse of $\phi$. Similarly to the proof technique of \cite{shalit2017estimating}, the following derivations shows the result.
\begin{align}
    \epsilon_{CF} - \epsilon_{F} &= \int_{\mathcal T}\int_{\mathcal X}l_{h,\phi}(\mathbf x, t)\cdot(p(\mathbf x)p(t)-p(\mathbf x,t))d\mathbf xdt\nonumber\\
    &= \int_{\mathcal T}\int_{\mathcal R}l_{h,\phi}(\psi(\mathbf r), t)\cdot(p(\psi(\mathbf r))p(t)-p(\psi(\mathbf r),t))d\psi(\mathbf r)dt\nonumber\\
    &= \int_{\mathcal T}\int_{\mathcal R}l_{h,\phi}(\psi(\mathbf r), t)\cdot(p_{\phi}(\mathbf r)p_{\phi}(t)-p_{\phi}(\mathbf r,t))J_{\psi}J_{\psi}^{-1}d\mathbf rdt\nonumber\\
    &\leq B_{\phi}\cdot \sup_{g\in G} \Big|\int_{\mathcal T}\int_{\mathcal R} g(\mathbf r,t)\cdot (p_{\phi}(\mathbf r)p_{\phi}(t) - p_{\phi}(\mathbf r,t))d\mathbf rdt \Big|.\nonumber
\end{align}
For the 2nd equality, the distribution $p_{\phi}$ over $\mathcal R \times \mathcal T$ can be obtained by the standard change of variables formula, using the determinant of the Jacobian of $\psi(\mathbf r)$, denoted $J_\psi$ giving $p(\psi(\mathbf r),t)=p_{\phi}(\mathbf r,t)J_\psi$ (which cancels with the inverse Jacobian that appears after the change of variables in the differential term). The last inequality comes from the assumption that $l_{h,\phi}(\mathbf x, t)/B_{\phi}\in G$, which is justified and extensively discussed in \cite{shalit2017estimating}.

\textbf{Theorem 2} (Generalization bound for selected treatment tuples $t_1$ and $t_2$). \textit{With the notation introduced above, it holds that,
\begin{align}
    \epsilon_{(t_1,t_2)}(f)/2 &\leq  
    \epsilon_{F}(t_1) + \sup_{g\in G} \Big|\int_{\mathcal R} g(\mathbf r)\cdot (p_{\phi}(\mathbf r) - p_{\phi}(\mathbf r|t_1))d\mathbf r \Big|  
    +\epsilon_{F}(t_2) \\
    &+ \sup_{g\in G} \Big|\int_{\mathcal R} g(\mathbf r)\cdot (p_{\phi}(\mathbf r) - p_{\phi}(\mathbf r|t_2))d\mathbf r \Big| - \sigma_{Y(t_1)} - \sigma_{Y(t_2)},
\end{align}
where $\sigma_{Y(t_1)}$ and $\sigma_{Y(t_2)}$ stand for the variance of the random variables $Y(t_1)$ and $Y(t_2)$, respectively, under the distribution $p(\mathbf x)$.}

\textit{Proof.}
\begin{align}
    \epsilon_{(t_1,t_2)}(f) &= \int_{\mathcal X}(f(t_1,x) - m(t_1,x) + m(t_2, x) - f(t_2,x))^2 p(x)dx \\
    &\leq 2\int_{\mathcal X}(f(t_1,x) - m(t_1,x))^2 p(x)dx + 2\int_{\mathcal X}(f(t_2,x) - m(t_2,x))^2 p(x)dx\\
    & = 2(\epsilon_{CF}(t_1) - \sigma_{Y(t_1)}(p(x))) + 2(\epsilon_{CF}(t_2) - \sigma_{Y(t_2)}(p(x)))\\
    &\leq 2(\epsilon_{F}(t_1) + \sup_{g\in G} \Big|\int_{\mathcal R} g(\mathbf r)\cdot (p_{\phi}(\mathbf r) - p_{\phi}(\mathbf r|t_1))d\mathbf r \Big|   - \sigma_{Y(t_1)}(p(x))) \\
    &+2(\epsilon_{F}(t_2) + \sup_{g\in G} \Big|\int_{\mathcal R} g(\mathbf r)\cdot (p_{\phi}(\mathbf r) - p_{\phi}(\mathbf r|t_2))d\mathbf r \Big|   - \sigma_{Y(t_2)}(p(x))).
\end{align}
The first inequality holds by the fact that $(a + b)^2\leq 2a^2 + 2b^2$ for any $a,b\in\mathbb R$. The second equality holds by Lemma 2, given below, and the last inequality holds by the same arguments as those used in Theorem 1.

\textbf{Lemma 2.} \textit{The following derivation holds.}
\begin{align}
    \epsilon_{CF}(t) &= \int_{\mathcal X}l_{h,\phi}(x, t)p(x) dx\\
    &= \int_{\mathcal X}\int_{\mathcal Y}(Y(t)-f(x,t))^2 p(Y(t)|x)p(x) dY(t)dx\\
    &= \int_{\mathcal X}\int_{\mathcal Y}(f(x,t)-m(x,t))^2 p(Y(t)|x)p(x) dY(t)dx \\
    &+ \int_{\mathcal X}\int_{\mathcal Y}(m(x,t) - Y(t))^2 p(Y(t)|x)p(x) dY(t)dx \\
    &+ 2\int_{\mathcal X}\int_{\mathcal Y}(f(x,t)-m(x,t))(m(x,t) - Y(t)) p(Y(t)|x)p(x) dY(t)dx\\
    &= \int_{\mathcal X}(f(x,t)-m(x,t))^2 p(x) dx + \sigma_{Y(t)}(p(x)).
\end{align}
The third term in the third equality evaluates to zero because $m(x,t):=\int_{\mathcal Y}Y(t)p(Y(t)|x)$ and we have defined the variance of $Y(t)$ with respect to the distribution $p(x)$ as $\sigma_{Y(t)}(p(x)):=\int_{\mathcal X}\int_{\mathcal Y}(m(x,t) - Y(t))^2 p(Y(t)|x)p(x) dY(t)dx$.

\section{Data generating functions for experiments} \label{sec:ap_data}
This section describes the data generating mechanisms used in our experiments.

\paragraph{Synthetic.}
We generate synthetic data similar to \cite{nie2021vcnet}.
With covariates $\mathbf{x} \in \mathbb{R}^6$ all drawn from a uniform distribution between 0 and 1, we generate the continuous dosages and outcomes as follows,
\begin{align}
    \tilde{s}| \mathbf{x} =&  \frac{10 \sin( \max(x_1, x_2, x_3)) + \max(x_3, x_4, x_5)^3}{1 + (x_1 + x_5)^2} + \sin (0.5 x_3)(1 + \exp(x_4 - 0.5 x_3 )) + \\ & x_3^2 + 2 \sin(x_4) + 2x_5 - 6.5 + \mathcal{N}(0, 0.25), \nonumber \\
    y| \mathbf{x}, s =& \cos(2 \pi(s - 0.5))  \left( s^2 + \frac{4 \max(x_1,x_6)^3}{1 + 2x_3^2} \sin(x_4) \right)  + \mathcal{N}(0, 0.25), 
\end{align}
where $ s = (1 + \exp(-\tilde{s}))^{-1}$.

\paragraph{IHDP Continuous.}
The IHDP dataset contains 25 covariates with binary treatments and continuous outcomes \cite{hill2011bayesian}.
Disregarding the treatments and outcomes, we use the covariates to generate continuous dosages and treatments to test our method.
We follow the data generating procedure of \cite{nie2021vcnet}.
The dosages are generated according to,
\begin{align}
    \tilde{s}| \mathbf{x} &=  \frac{2x_1}{1 + x_2} + \frac{2 \max(x_3, x_5, x_6)}{0.2 + \min (x_3, x_5, x_6)} + 2 \tanh \left(5\frac{ \sum_{i\in I} x_i - c_2}{ |I|} \right) - 4 + \mathcal{N}(0, 0.25), \\
    y| \mathbf{x}, s &= \frac{\sin(3 \pi s)}{1.2 - s} \left( \tanh \left(5\frac{\sum_{i\in J} x_i - c_1}{|J|} \right) + \exp \left( \frac{0.2 (x_1 - x_6)}{0.5 + \min(x_2, x_3, x_5)} \right) \right)  + \mathcal{N}(0, 0.25), \\
    c_1 &= \mathbb{E}_{p(\mathbf{x})}\left[\frac{\sum_{i\in J}x_i}{|J|} \right], \\
    c_2 &= \mathbb{E}_{p(\mathbf{x})}\left[\frac{\sum_{i\in I}x_i}{|I|} \right],
\end{align}
where $ s = (1 + \exp(-\tilde{s}))^{-1}$, $I = \{16,17,18,19,20,21,22,23,24,25 \}$, and $J = \{4, 7,8,9,10,11,12,13,14,15 \}$.

\paragraph{News.}
This dataset contains words sampled from 5000 news articles \cite{newman2008bag}.
The covariates are word counts.
We generated continuous dosage and outcomes by following the data generation method listed in \cite{bica2020estimating}.
We first sample three vectors $\mathbf{v}'_i \sim \mathcal{N}(0,1)$, with $\mathbf{v}_i = \mathbf{v}'_i / || \mathbf{v}'_i||_2$ for $i = 1,2,3$. 
The dosages are drawn from a distribution $s \sim \text{Beta}(\alpha, s_w)$, where we set $\alpha = 2$ and, 
\begin{align}
    s_w= \max \left(1, \left|\frac{2\mathbf{x}^T \mathbf{v}_2 }{ \mathbf{x}^T \mathbf{v}_1} \right| \right).
\end{align}
Now we sample the outcomes according to,
\begin{align}
    y' &= \exp \left( \left|\frac{\mathbf{x}^T \mathbf{v}_2 }{ \mathbf{x}^T \mathbf{v}_1} \right| - 0.3 \right) \\
    y &= 2 \left( \max( -2, \min(2, y' ) + 20 \mathbf{x}^T \mathbf{v}_3 * (4(s - 0.5)^2 ) * \sin \left(\frac{\pi s}{2} \right) \right) + \mathcal{N}(0, 0.25).
\end{align}

\section{Algorithmic details}
\label{sec:alg_details}

\subsection{Architectures and training details}
For the VCNet and DRNet architectures, we use 2 layers each for the representation part of the network and for the regression part. Each layer has 50 hidden unit with ELU activations. Following \cite{nie2021vcnet}, we use B-spline with degree 2 and knots placed at $\{1/3, 2/3\}$ for VCNet and 5 regression heads for DRNet. For the MLP model, we use a 4-layers network to represent similar power of approximations to ensure fair comparison. We optimise the networks using Adam \cite{kingma2014adam} with a weight decay of 0.005 for regularisation and a batch size of 1000. Learning rate is chosen within the set $\{0.01, 0.005, 0.001, 0.0005, 0.0001, 0.00005 \}$ using the procedure outlined in \ref{subsec:hyperp}. Each data set is split into a train/validation/test set with ratios 0.6/0.2/0.2.
To avoid overfitting, we stop the training if the validation loss did not improve after 50 epochs.

\paragraph{Propensity score regularization (-PS methods)} In addition to the  representation net $\phi: \mathcal{X} \rightarrow \mathbb{R}$ and to the prediction net $h_1: \mathcal{R} \times \mathbb{R} \rightarrow \mathcal{Y}$, propensity score regularized methods also include a separate head $h_2:\mathcal R \rightarrow \mathcal{T}$. Parameters are tuned by minimizing the loss
\begin{align}
    \mathcal{L}_{PS}(\phi, h_1, h_2) = \frac{1}{n}\sum_{i=1}^n (y_i - h_1(\phi(\mathbf x_i),t_i))^2  +\alpha \cdot\text{CrossEntropy}(h_2(\phi(\mathbf x_i)), t_i).
\end{align}
Average treatment effects are then estimated by considering an additional perturbation term following \cite{shi2019adapting} and \cite{nie2021vcnet}. $\alpha$ is treated as a hyperparameter and chosen within the set $\{0.5, 1 \}$ using the procedure detailed in \ref{subsec:hyperp}. 
The implementation in practice follows the publicly available code of \cite{nie2021vcnet}.

\paragraph{Targeted regularization (-TR methods)} Methods labeled -TR use the functional targeted regularization approach presented in \cite{nie2021vcnet} which optimizes the loss function
\begin{align}
\mathcal{L}_{TR}(\phi, h_1, h_2, \epsilon_n)=  \mathcal{L}_{PS}(\phi, h_1, h_2)  + \frac{\beta}{n} \sum_{i=1}^n \left(y_i - h_1(\phi(\mathbf x_i),t_i) - \frac{\epsilon_n(t_i)}{h_2(\phi(\mathbf x_i))} \right)^2
\end{align}
where $\epsilon_n(\cdot)=\sum_{k=1}^{K_n} a_k \psi_k(\cdot)$ is modelled through $K_n$ spline basis functions $\psi_k$ of degree 2. Following \cite{nie2021vcnet}, we select the learning rate for $\epsilon_n(\cdot)$, $\beta$ and the number of spline knots within the sets \{0.001, 0.0001\}$, \{20, 10, 5 \}/ \sqrt{n}$ and $\{5,10,20\}$, respectively. Again, the implementation in practice follows the publicly available code of \cite{nie2021vcnet}.

\paragraph{IPM regularization (-HSIC and -Wass methods)}
IPM regularized methods minimize the proposed loss in Equation (\ref{objective}) in the main body of this paper, where the $\gamma$ is selected within the set $\{10^{i / 6},\ i=-18,-17 , \cdots, 9, 10\}$ using the procedure in \ref{subsec:hyperp}.

\paragraph{Generalised Propensity Score (GPS)} We use the linear implementation from \cite{hirano2004propensity}.

\subsection{Hyper-parameters tuning}
\label{subsec:hyperp}
We use grid-search to tune the hyper-parameters. Namely, we generate a dataset for each hyperparameters setting, randomly splitting it into a train/test set with a ratio of 0.8/0.2 and we choose the hyperparameters values giving the best MISE test score.

\end{document}